\documentclass[runningheads]{llncs}

\usepackage{graphicx}
\usepackage{amsmath}
\usepackage{amssymb}
\usepackage{hyperref}
\usepackage{color}

\usepackage[ruled,vlined,linesnumbered]{algorithm2e}
\usepackage{algpseudocode}
\usepackage{booktabs}
\usepackage{todonotes}
\usepackage{mathtools}
\usepackage{units}
\usepackage{comment}
\usepackage{marvosym}
\usepackage{tikz}
\usepackage{multirow}
\usepackage{array}

\iffalse
\newcommand{\vw}[1]{}
\newcommand{\luca}[1]{}
\newcommand{\MB}[1]{}
\newcommand{\GG}[1]{}
\renewcommand{\todo}[1]{}
\else
\newcommand{\vw}[1]{\todo{\emph{VW}: #1}}
\newcommand{\luca}[1]{\todo{\emph{LB}: #1}}
\newcommand{\GG}[1]{\todo{\emph{GG}: #1}}
\newcommand{\MB}[1]{\todo{\emph{MB}: #1}}
\fi

\newcommand{\PreserveBackslash}[1]{\let\temp=\\#1\let\\=\temp}
\newcolumntype{C}[1]{>{\PreserveBackslash\centering}p{#1}}
\newcolumntype{R}[1]{>{\PreserveBackslash\raggedleft}p{#1}}
\newcolumntype{L}[1]{>{\PreserveBackslash\raggedright}p{#1}}

\spnewtheorem{model}[theorem]{Model}{\bfseries}{\itshape }

\begin{document}

\title{Abstraction-Guided Truncations for Stationary Distributions of Markov Population Models}
\titlerunning{Abstraction-Guided Truncations for Stationary Distributions of MPMs}

\author{Michael Backenköhler\inst{1}\textsuperscript{,\Letter}, Luca Bortolussi\inst{2,3}, Gerrit Großmann\inst{1}, Verena Wolf\inst{1,3}}
\authorrunning{M.\ Backenköhler et al.}
\institute{
${}^1$Saarbrücken Graduate School of Computer Science,
Saarland University,
Saarland Informatics Campus E1 3,
Saarbrücken, Germany\\
\textsuperscript{\Letter} \email{michael.backenkoehler@uni-saarland.de}\\
${}^2$Univeristy of Trieste, Trieste, Italy\\
${}^3$
Saarland University,
Saarland Informatics Campus E1 3,
Saarbrücken, Germany\\
}
\maketitle
\begin{abstract}
To understand the  long-run behavior of Markov population models, the computation of the stationary distribution    is often a crucial  part.
We propose a truncation-based approximation that employs a
state-space lumping scheme, aggregating states in a grid structure.
The resulting approximate stationary distribution is used to iteratively refine relevant
and
truncate irrelevant parts of the state-space.
This way, the algorithm learns a well-justified finite-state
projection tailored to the stationary behavior.
We demonstrate the method's applicability to a wide range of non-linear problems  with complex stationary behaviors.
\end{abstract}
\section{Introduction}
In many areas of science, stochastic models  of interacting populations can describe systems in which the discrete population sizes evolve stochastically in continuous time.
Such problems naturally occur in a wide range of areas such as chemistry~\cite{gillespie1977exact}, systems biology~\cite{wilkinson2018stochastic,ullah2011stochastic}, epidemiology~\cite{mode2000stochastic} as well as    queuing systems~\cite{breuer2003markov} and finance~\cite{pardoux2008markov}.

Interactions between agents, commonly referred to as \emph{reactions}, happen at exponentially distributed random times. 
Their rate depends on the current system state, i.e.\ the population sizes.
This results in a continuous-time Markov chain semantics~\cite{anderson2012continuous}.
An important part of the analysis of such models concerns their long-run behavior.
Given an ergodic underlying Markov chain, the chain's stationary distribution characterizes this behavior.
For some special model classes, such as zero-deficiency networks~\cite{anderson2011continuous}, analytical solutions for the stationary distribution are known.
However, most models require numerical approaches, often based on some form of approximation to guarantee tractability.
Those approaches can be based on stochastic simulation~\cite{gillespie1977exact} (which for steady-state analysis tends to be slow and inaccurate) or moment-bounds via mathematical programming~\cite{kuntz2017rigorous}.
Here, we draw on numerical approaches based on state-space truncation, which represent a viable option to approximate stationary distributions~\cite{kuntz2021approximations}.
Truncation-based approaches have the benefit of describing the complete dynamics within a finite subset of the typically very large or infinite state-space.
As such, they enable the approximation of complex distributions that are not well-described by low-order moments.


The main step in the computation of such an approximation is the identification of a suitable truncation,
i.e.\ a subset of the state-space encompassing most of the stationary probability mass.
Existing methods typically rely on Foster-Lyapunov drift conditions to define such subsets~\cite{dayar2011bounding}.
While these truncations come with bounds on the contained stationary probability mass, they typically are far larger than necessary.
The truncation is usually strongly constrained by the form of the chosen Lyapunov function~\cite{gupta2014scalable,dayar2011bounding}.
Optimizing over possible functions to identify efficient truncations is technically challenging and, to our knowledge, has not been demonstrated for general reaction networks~\cite{milias2014optimization}.

In this work, we address the identification of suitable truncations by using an aggregation-refinement scheme.
Initially, a Lyapunov analysis yields a set containing at least $1-\epsilon$ of the stationary probability mass.
On this subset of the state-space, we apply an aggregation scheme that groups together states in  hypercube macro-states.
Throughout each of these macro-states, we assume a uniform distribution among its constituent micro-states.
This allows us to roughly analyze large portions of the state-space with exponentially fewer variables.
We then iteratively truncate and refine the approximation based on the stationary distribution of this aggregated Markov chain. 
We keep only the most relevant macro-states and  continue this scheme until the macro-states contain a single original state. 
In this way, we arrive at an effective truncation to compute an approximation of the stationary distribution.

We investigate the approximation results on case studies with known stationary distributions and complex models with intricate stationary distributions.
We evaluate the truncation quality by assessing the stationary probability mass captured.
To this end, we use analytical solutions and bounds given by a Lyapunov analysis.
Further, we explore the control of the truncation size through the truncation parameter.
Finally, we demonstrate the method on the p53 oscillator model exhibiting a complex stationary distribution.

The rest of the paper is organized as follows: Section 2 discusses related work, Section 3 introduces background material, Section 4 is devoted to the description of our method, Section 5 presents an experimental validation, and finally Section 6 contains a final discussion. 

\section{Related Work}\label{sec:related}
For some specific models, analytical solutions for the stationary distribution have been found \cite{melykuti2014equilibrium,kurasov2018stochastic}. For the class of zero-deficiency networks, the stationary distribution is known to have a Poisson product form \cite{anderson2010product}. Monomolecular reaction networks can be solved explicitly, as well~\cite{jahnke2007solving}.

The analysis of countably infinite-sized state-spaces is often handled by pre-defined truncations~\cite{kwiatkowska2011prism}.
Sophisticated state-space truncations for the (uncon\-ditioned) forward analysis have been developed that give lower bounds.
They typically provide a trade-off between computational load and tightness of the bound~\cite{munsky2006finite,lapin2011shave,andreychenko2011parameter,henzinger2009sliding,mikeev2013fly}.
Such methods cannot be directly applied to the estimation of stationary distributions because the approximation usually introduces a sink-state.

Truncations for stationary distributions often involve re-direction schemes for transitions leaving and entering the subset.
A comprehensive survey of such state-space truncation methods can be found in \cite{kuntz2021stationary}.
A popular method of identifying truncations is the construction of a suitable Lyapunov function.
Beyond their use for establishing ergodicity \cite{meyn1993stability,gupta2014scalable,dayar2011bounding},
these functions can be used to obtain truncations, guaranteed to contain a certain amount of stationary probability mass \cite{dayar2011bounding}.
Using Lyapunov functions for the construction of truncations often leads to very conservative sets \cite{milias2014optimization}.
Different approaches have been employed to find truncations:
In \cite{gupta2017finite} SSA estimates are used to set up an increasing family of truncations.

Apart from approaches based on state-space truncations, moment-based approaches have been particularly popular recently~\cite{ghusinga2017exact,dowdy2018bounds,kuntz2017rigorous,sakurai2017convex}.
Such approaches are based on the fact that particular matrices of distributional moments such as mean and variance are positive semi-definite.
Along with linear constraints stemming from the Kolmogorov equations \cite{backenkohler2016generalized}, a semi-definite program can be formulated and solved using existing tools.
While this method is suited to compute bounds on both moments and subsets of the state-space, its application is limited, due to numerical issues inherent in the formulation \cite{dowdy2018bounds}.

An approach where quantities are only described in terms of their magnitude has been proposed in \cite{ceska2019semi}. This allows for an efficient qualitative analysis of both dynamic and transient behavior.

An aggregation scheme similar to the one used here has been previously proposed in \cite{backenkohler2020analysis} to analyze the bridging problem on Markovian population models.
This is the problem of analyzing process dynamics under both initial and terminal constraints.

Aggregation-based numerical methods for computing the stationary distribution 
of discrete or continuous-time Markov chains have been studied in
previous work. Popular approaches rely on an alternation of aggregation and 
disaggregation of the state-space \cite{stewart1994introduction,schweitzer1991survey}.
In the case of stiff chains, such aggregations are typically based on 
a separation of time-scales \cite{cao1985iterative}.
However, these methods have been developed for finite chains with arbitrary structure and are motivated by numerical issues of standard methods such as 
the power method or Jacobi iteration \cite{stewart1994introduction}.
They do not consider a truncation of irrelevant states, while
here our aggregation approach is used to determine the most relevant states
under stationary conditions in large or infinite chains with population structure.

\section{Preliminaries}\label{sec:prelim}
\subsection{Markovian Population Models}
A Markovian population model (MPM)
describes the stochastic interactions
among agents of distinct types in a well-stirred system.
This assumes that all agents are equally distributed in space, which
allows us to keep track only of the overall copy number of agents for each type.
Therefore the state-space is $\mathcal{S}\subseteq\mathbb{N}^{n_S}$ where
$n_S$ denotes the number of agent types or populations.
Interactions between agents are expressed as \emph{reactions}.
These reactions have associated
gains and losses of agents, given by non-negative integer vectors   
${v}_j^{-}$ and ${v}_j^{+}$ for reaction $j$, respectively. The overall change by a reaction is given by the vector $v_j = v_j^+ - v_j^-$.
A reaction between agents of types $S_1,\dots, S_{n_S}$ is specified in the following form:
\begin{equation}\label{eq:reaction}
    \sum_{\ell=1}^{n_S} v_{j\ell}^{-} S_\ell
    \xrightarrow{\alpha_j( x)}
    \sum_{\ell=1}^{n_S} v_{j\ell}^{+} S_\ell\,.
\end{equation}
The propensity function $\alpha_j$ gives the rate of the exponentially distributed firing
time of the reaction as a function of the current system state $x\in \mathcal{S}$.
In population models, \emph{mass-action} propensities are most common.
In this case the firing rate is given by the product of the number
of reactant combinations in $x$ and a
\emph{rate constant} $c_j$, i.e.
\begin{equation}\label{eq:stoch_mass_action}
    \alpha_j({x})\coloneqq c_j\prod_{\ell=1}^{n_S}\binom{x_\ell}{v_{j\ell}^{-}}\,.
\end{equation}
In this case, we give the rate constant in \eqref{eq:reaction} instead of the function $\alpha_j$.
For a given set of $n_R$ reactions, we define a stochastic
process $\{{{X}}_t\}_{t\geq 0}$ describing the evolution of the population
sizes over time $t$.
Due to the assumption of exponentially distributed firing times\footnote{Note that in addition mild regularity assumptions
are   necessary for the existence of a unique CTMC $X$, such as non-explosiveness \cite{anderson2012continuous}.
These assumptions  are  typically
valid for realistic reaction networks.},  $ X$ is
a continuous-time
Markov chain (CTMC) on $\mathcal{S}$ with infinitesimal  generator matrix $Q$, where
the entries of $Q$ are
\begin{equation}\label{eq:cme_generator}
    Q_{ x,  y} = \begin{cases}
        \sum_{j: x+ v_j = y}\alpha_j( x)\,,&\text{if}\; x\neq
         y,\\[1ex]
        -\sum_{j=1}^{n_R} \alpha_j( x)\,, &\text{otherwise.}
    \end{cases}
\end{equation}
The probability distribution over time is given by an
initial value problem.
Given an initial state $x_0$, the distribution\footnote{In the sequel, 
we assume an enumeration of all states in  $\mathcal{S}$. We simply write $x_i$ for the state with index $i$ and drop this notation for entries of a state $x$.  }
\begin{equation}\label{eq:forw_prob}
\pi(x_i, t)=\Pr(X_t=x_i\mid X_0=x_0),\quad t\geq 0
\end{equation}
evolves according to the Kolmogorov forward equation
\begin{equation}\label{eq:forward}
\frac{d}{dt}\pi(t) = \pi(t) Q\,,
\end{equation}
where $\pi(t)$ is an arbitrary vectorization $(\pi(x_1,t), \pi(x_2,t),\dots,\pi(x_{|\mathcal{S}|},t))$ of the states.

\paragraph{Example.} Consider a birth-death process as a simple example. This model is used to describe a wide variety of phenomena and often constitutes a sub-module of larger models.
For example, it represents an M/M/1 queue with service rates being linearly dependent on the queue length.
Note that even for this simple model, the state-space is countably infinite.
\begin{model}[Birth-Death Process]\label{model:bd}
The model consists of exponentially distributed arrivals and service times proportional to queue length. It can be expressed using two mass-action reactions:
$$ \varnothing \xrightarrow{\mu} S \qquad\text{and}\qquad S \xrightarrow{\gamma} \varnothing\,.$$
The initial condition $X_0=0$ holds with probability one.
\end{model}

\subsection{Stationary Distribution}\label{sec:stationary_dist}
Assuming
ergodicity  
of the underlying chain, a stationary distribution $\pi_{\infty}$ is an invariant distribution, namely a fixed point of the Kolmogorov forward equation \eqref{eq:forward}.
Let $\pi_{\infty}$ be the vector description of a stationary distribution. It then  satisfies
\begin{equation}\label{eq:stationary}
0=\pi_{\infty}Q\quad\text{and}\quad 1=\sum_{x\in\mathcal{S}}\pi_{\infty}(x)
\end{equation}
as a fixed point of the Kolmogorov equation \eqref{eq:forward}.
Stationary distributions are connected to the \emph{long-run} behavior of an MPM \cite{dayar2011bounding}, as the system's distribution will converge to the (unique)
stationary distribution.
The connection of the stationary distribution to the long-run behavior becomes clear when considering the ergodic theorem. 
For some $A\subseteq\mathcal{S}$,
\begin{equation}\label{eq:ergodic}
    \lim_{T\to\infty}\frac{1}{T}\int_0^T 1_A(X_t)\,dt
    = \sum_{x\in A}\pi_{\infty}(x)\,.
\end{equation}
Thus, the mean occupation time for set $A$ over infinite trajectories is the stationary measure for $A$.
Eq.~\eqref{eq:ergodic} shows that we can assess long-run behavior using the stationary distribution and vice-versa.

\paragraph{Example.} Returning to the example of Model~\ref{model:bd} it is obvious that the state-space is irreducible.
Further, we can easily show, that the stationary distribution is Poissonian with rate $\mu/\gamma$:
$$ \pi_{\infty}(x)=\frac{{(\mu/\gamma)}^{x}\exp(-\mu/\gamma)}{x!}\,.$$

For simplicity, we assume throughout that the state-space is composed of a single communicating class.
Checking ergodicity given a countably infinite number of states is achieved by providing a suitable Foster-Lyapunov function \cite{meyn2012markov}.
Some automated techniques have been proposed for this task \cite{dayar2011bounding,gupta2014scalable,milias2014optimization}.

\subsection{Truncation-Based Approximation of $\pi_{\infty}$}\label{sec:fsp}
\begin{figure}[t]
    \centering
    \begin{minipage}{0.6\textwidth}
    \includegraphics[width=\textwidth]{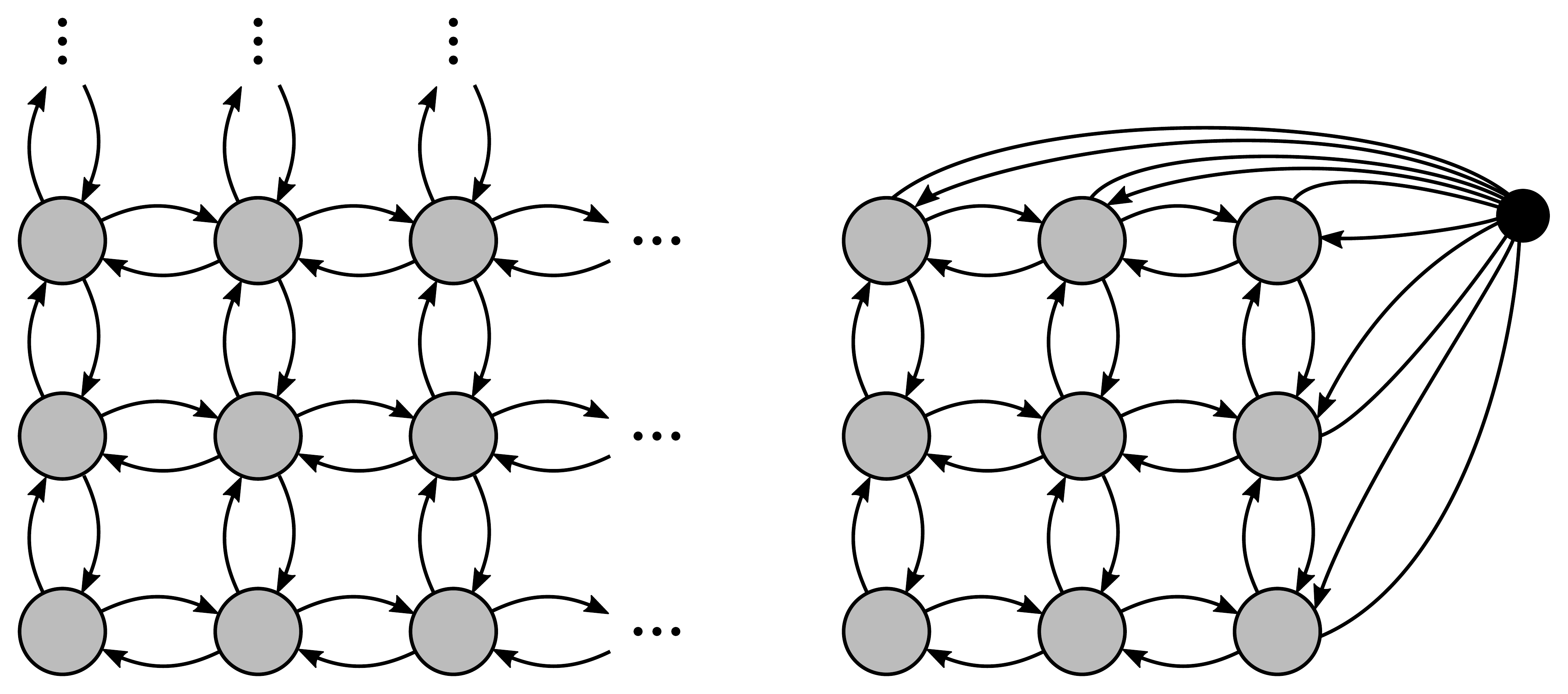}
    \end{minipage}
    \begin{minipage}{0.35\textwidth}
    \includegraphics[width=\textwidth]{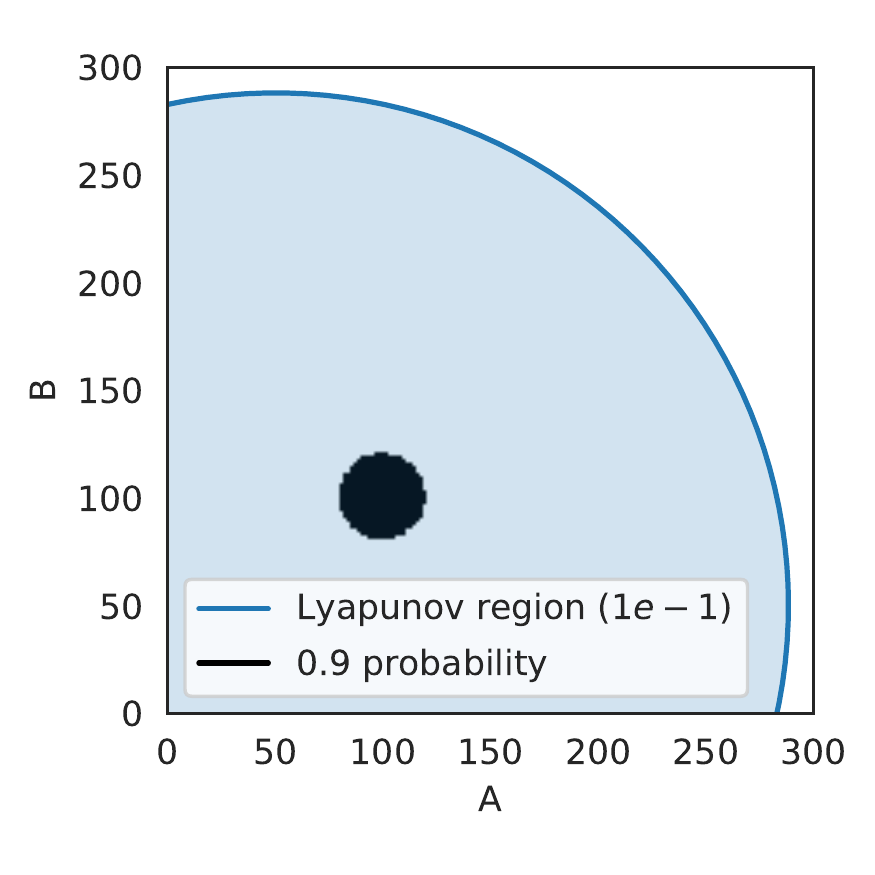}
    \end{minipage}
    \caption{(left) A countably infinite state-space. (center) Outgoing transitions are re-directed (according to the reentry distribution) to states that have incoming transitions from outside the truncation. (right) A comparison of the area perscribed by a Lyapunov analysis using Geobound and threshold 0.1 and the minimal area containing 0.9 stationary probability mass. The model is a parallel birth death process (Model~\ref{model:par_bd}).}
    \label{fig:truncation}
\end{figure}

In many relevant cases, the state-space is huge or infinite and therefore the stationary solution cannot be computed directly.
To make such a computation possible we have to restrict ourselves to a finite manageable subset of the state-space and assume the majority of the probability mass is concentrated within that finite subset.
The main problem is to deal with the transitions leading to and from the truncated set (cf.\ Figure~\ref{fig:truncation}).
In forward analysis, the outgoing transitions are simply redirected into a sink-state.
This way, a forward analysis provides lower bounds since mass leaving the truncation does not re-enter.
This approach, however, is unsuitable for the computation of stationary distributions because mass would accumulate in the sink-state leading to a distribution assigning all mass to it.
Therefore, transitions leaving the truncation need to be redirected back into the truncation.

The process' dynamics outside the truncation are defined by the \emph{stochastic complement} \cite{spieler2014numerical}.
If its behavior was known, one could redirect outgoing to incoming transitions optimally and preserve the correct stationary distribution.
However, this reentry distribution is typically unknown in most relevant cases.
Many different reentry distributions have been used, such as redirecting to some internal state or states with incoming transition from outside the truncation.
Reference~\cite{kuntz2021approximations} provides a comprehensive review of such methods.

The most natural choice is to pick a reentry distribution that redirects mass to states with incoming transitions from truncated states (cf.\ Figure~\ref{fig:truncation} (center)).

Using varying redirections, we can compute bounds on the stationary probability conditioned on a truncation \cite[(Thm. 14)]{spieler2014numerical}.
To do this, one has to compute the stationary distribution for every possible way of connecting all outgoing to a single incoming transition.
Naturally, such an algorithm is rather expensive since one has to solve a linear system for each combination.
Therefore this method of computing bounds is costly on very large truncations, often given by Lyapunov functions.

When computing an approximation instead of bounds, we employ a uniform redirection scheme:
Outgoing transitions are split uniformly among incoming transitions.
Due to the threshold-based truncation scheme, we are likely to end up with a somewhat uniform distribution over in-boundary states (see Section~\ref{sec:alg}).

The identification of good truncations remains a major task in such approximations.
Using approaches such as Lyapunov functions (Section~\ref{sec:lyapunov}) \cite{dayar2011bounding} or moment-bounds \cite{kuntz2021approximations} can provide a good initial estimate, but typically the resulting truncations are far larger than necessary.
This leads to dramatically increased computational costs, especially when bounding methods mentioned above are performed.
Until a system for a larger truncation is solved, the precise location of  most of the probability mass is often unknown.
Instead of solving the full system for such a large space, we employ an aggregation scheme to cover large areas of the state-space with exponentially fewer variables.

Error bounds have been derived for increasing  truncation sets 
in the case of linear Lyapunov functions  \cite{gupta2017finite}.
However, until now it has not been shown that these bounds are applicable in practice \cite{meyn1994computable}.
Alternatively, one can monitor the product of the probability-ouflow rate and the maximum L1-norm, which bounds the approximation error up to a constant $M>0$, assuming a linear Lyapunov function exists \cite{gupta2017finite}.

\subsection{Lyapunov Bounds}\label{sec:lyapunov}
It is well-known that for a CTMC $X$, ergodicity can be proven by a Lyapunov function $g:\mathcal{S}\to\mathbb{R}_+$ \cite{meyn1993stability,dayar2011bounding}.
Given the $g$, we define its \emph{drift} $d$ as its average infinitesimal change, which is obtained applying the generator $Q$ to $g$. 
\begin{equation}
    d(x) = \sum_{j=1}^{n_R} \alpha_j(x) (g(x+v_j) -  g(x))
\end{equation}
Usually, such a function $g$ grows in all directions on the positive orthant, while its drift $d(x)$ decreases in all directions.
More formally, $g$ is characterized by having finite level sets $\{x\in\mathcal{S} \mid g(x) < l\}$ for all $l > 0$.
At the same time,
\begin{equation}\label{eq:lyapunov_set}
    \mathcal{C}_{\epsilon_{\ell}} = \{ x\in\mathcal{S} \mid
    \frac{\epsilon_{\ell}}{c}d(x) > \epsilon_{\ell} - 1\}
\end{equation}
should be finite, where $\infty> c\geq \sup_{x\in\mathcal{S}} d(x)$.
In this case, $\mathcal{C}_{\epsilon_{\ell}}$ contains at least $1-\epsilon_{\ell}$ of stationary probability mass for any $\epsilon_{\ell}\in(0,1)$ \cite[Thm.~8]{spieler2014numerical}.
Given that $\mathcal{C}_{\epsilon_{\ell}}$ is finite, the chain is ergodic and
\begin{equation}
    \sum_{x\in\mathcal{C}_{\epsilon_{\ell}}}\pi(x)> 1 - \epsilon_{\ell}
\end{equation}
bounding the stationary probability mass contained within $\mathcal{C}_{\epsilon_{\ell}}$.

In many cases, simple choices of $g$ such as the L1- or L2- norm are sufficient.
However, the sets resulting from such functions are often very conservative.
Consider Figure~\ref{fig:truncation} (right) as an example, where the Lyapunov truncation with $\epsilon_{\ell}=0.1$ 
for two parallel birth death processes (Model~\ref{model:par_bd}) is compared to the smallest set containing 0.9 of stationary probability.
Clearly, the area given by the Lyapunov function is magnitudes larger than necessary to capture probability mass consistent with $\epsilon_{\ell}$.

We employ this approach to both identify initial truncations and estimate errors in the evaluation.
Specifically, we employ the tool Geobound\footnote{\url{https://mosi.uni-saarland.de/tools/geobound}} with L2-norm as function $g$ implementing techniques presented in \cite{dayar2011bounding}.

\section{Method}\label{sec:method}
In this work, we propose a  method to identify a truncation that optimizes the trade-off between the size of the considered state-space and the approximation error due to the finite state-space projection.
To this end, we start with a very coarse-grained model abstraction that we refine iteratively. 
The coarse-grained model is based on an grid-shaped aggregation (i.e., lumping) scheme that identifies a set of macro-states.
These macro-states can be used to compute an interim model solution that guides the refinement in the next step.
We perform refinements until the approximation arrives at the resolution of the original model (i.e., each macro-state has only one constituent) such that the aggregation introduces no approximation error.

We explain the construction of macro-states in Section~\ref{sec:aggregation} and their initialization in Section~\ref{sec:initagg}.
We present the iterative refinement algorithm in Section~\ref{sec:alg}.

\subsection{State-Space Aggregation}\label{sec:aggregation}
A macro-state is a collection of micro-states (or simply states) treated as one state in the aggregated model, which can be seen as an abstraction of the original model.
The aggregation scheme defines a partitioning of the state-space.
We choose a scheme based on a grid structure. That is, each macro-state is a hypercube in $\mathbb{Z}_{\geq 0}^{n_S}$.

Hence, each macro-state $\bar{x}_i(\ell^{(i)},u^{(i)})$ (denoted by $\bar{x}_i$ for notational ease) can be identified using two vectors $\ell^{(i)}$
and $u^{(i)}$.
The vector $\ell^{(i)}$ gives the corner closest to the origin, while $u^{(i)}$
gives the corner farthest from the origin.
Formally,
\begin{equation}\label{eq:macros_state}
    \bar{x}_i = \bar{x}_i(\ell^{(i)},u^{(i)}) =  \{x\in\mathbb{N}^{n_S} \mid  \ell^{(i)}  \leq x  \leq u^{(i)} \},
\end{equation}
where '$\leq$' denotes element-wise comparison.

In order to solve the aggregated model, we need to define transition rates between macro-states.
Therefore, we assume that, given that the system is in a particular macro-state, all constituent states are equally likely (uniformity assumption).
This assumption is the reason why the aggregated model provides only a coarse-grained approximation. 

The uniformity assumption is a modeling choice yielding significant advantages.
Firstly, it eases the computation of the rates between macro-states and, therefore, makes a fast solution of the aggregated model possible.
Secondly, even though it induces an approximation error, it provides suitable guidance as uniformity assumption spreads out the probability mass conservatively.
Hence, it becomes less likely that regions of interest are disregard.
Lastly, the uniformity assumption is theoretically well-founded, as it stems from the maximum entropy principle: 
In the absence of concrete knowledge about the probability distribution inside a macro-state, we assume the distribution with the highest uncertainty, i.e., the uniform distribution. 

The grid structure makes the computation of transition rates between macro-states particularly convenient and computationally simple.
Mass-action reaction rates can be given in a closed-form,
due to the Faulhaber formulae~\cite{knuth1993johann} and more complicated rate functions such as Hill-functions can often be handled as well by taking appropriate integrals~\cite{backenkohler2020analysis}.

Suppose, we are interested in the transition rate from macro-state $\bar{x}_i$ to macro-state $\bar{x}_k$ according to reaction $j$.
Using the uniformity assumption, this is simply the mean rate of the states in $\bar{x}_i$ that go to $\bar{x}_k$ using $j$.
However, only a small subset of constituents in $\bar{x}_i$ are actually relevant for this transition.
Hence, we identify the subset of states of $\bar{x}_i$ that lie at the border to $\bar{x}_k$ and in such a way that applying reaction $j$ shifts them to a state in $\bar{x}_k$. Then, we sum up the corresponding rates of these states. Lastly, we normalize according to the number of states inside of $\bar{x}_i$.

It is easy to see that the relevant set of border states is itself an
interval-defined macro-state $\bar{x}_{i\xrightarrow{j}k}$.
To compute this macro-state
we can simply shift $\bar{x}_i$ by $v_j$, take the intersection
with $\bar{x}_k$ and project this set back.
Formally,
\begin{equation}\label{eq:transition_set}
    \bar{x}_{i\xrightarrow{j}k} = ((\bar{x}_i + v_j) \cap \bar{x}_k) - v_j\,,
\end{equation}
where the additions are applied element-wise to all states
making up the macro-states.
For ease of notation, we also define a general exit state
\begin{equation}
    \bar{x}_{i\xrightarrow{j}} = ((\bar{x}_i + v_j) \setminus \bar{x}_i) - v_j.
\end{equation}
This state captures all micro-states inside $\bar{x}_i$ that can leave the state via reaction $j$.

This uniformity assumption gives rise to the following $Q$-matrix of the aggregated model:
\begin{equation}\label{eq:lumped_q}
    \bar{Q}_{ \bar{x}_i,  \bar{x}_k} = \begin{cases}
        \sum_{j=1}^{n_R}{\bar\alpha}_j\left(\bar{x}_{i\xrightarrow{j}k}\right)/\left|\bar{x}_i\right|\,,&\text{if}\; \bar{x}_i\neq \bar{x}_k\\[1ex]
        -\sum_{j=1}^{n_R}{\bar\alpha}_j\left(\bar{x}_{i\xrightarrow{j}}\right)/{\left|\bar{x}_i\right|}\,, &\text{otherwise}
    \end{cases}
\end{equation}
where 
\begin{equation}\label{eq:lumped_propfun}
    \bar{\alpha}_j({\bar{x}}) = \sum_{x\in \bar{x}} \alpha_j(x).
\end{equation}
is the sum of all rates belonging to reaction $j$ in $\bar{x}$..

Under the assumption of polynomial rates, as is the case for mass-action
systems, we can compute the sum of rates over this transition set
efficiently using Faulhaber's formula.
As an example consider the following mass-action reaction
$ 2 X \xrightarrow{c} \varnothing\,. $
For macro-state
$\bar{x} = \{0, \dots, n\}$
we can compute the corresponding lumped transition rate
$$\bar{\alpha}(\bar{x})
=\frac{c}{2}\sum_{i=1}^n i (i - 1)
=\frac{c}{2}\sum_{i=1}^n (i^2 - i)
=\frac{c}{2}\left(\frac{2n^3+3n^2+n}{6} - \frac{n^2 + n}{2}\right)$$
eliminating the explicit summation in the lumped propensity function.

\subsection{Initial Aggregation}\label{sec:initagg}
The initial aggregated space $\hat{\mathcal{S}}^{(0)}$ should encompass all regions of the state-space that could contain significant mass because states outside this initial area will not be refined.
In principle, multiple approaches could be used to identify such a region.
One possibility is the computation of moment bounds for the stationary distribution~\cite{ghusinga2017exact,dowdy2018bounds}.
Based on these bounds on expectations and covariances, an initial truncation could be fixed.
The approach we use here is to identify such a  region by a Lyapunov analysis \cite{dayar2011bounding}.
This way, we obtain a polynomial describing a semi-algebraic subset of the entire state-space containing $1-\epsilon_{\ell}$ of the mass, where $\epsilon_{\ell}>0$ can be fixed arbitrarily.
These sets usually are far larger than a minimal set containing $1-\epsilon_{\ell}$ of stationary probability mass would be.
As an initial aggregation, we build an aggregation on a subset $[0..n]^{n_S}\subset \mathcal{S}$ containing the set prescribed by the Lyapunov analysis.

\subsection{Iterative Refinement Algorithm}\label{sec:alg}
\begin{figure}[t]
    \centering
    \includegraphics[scale=.6]{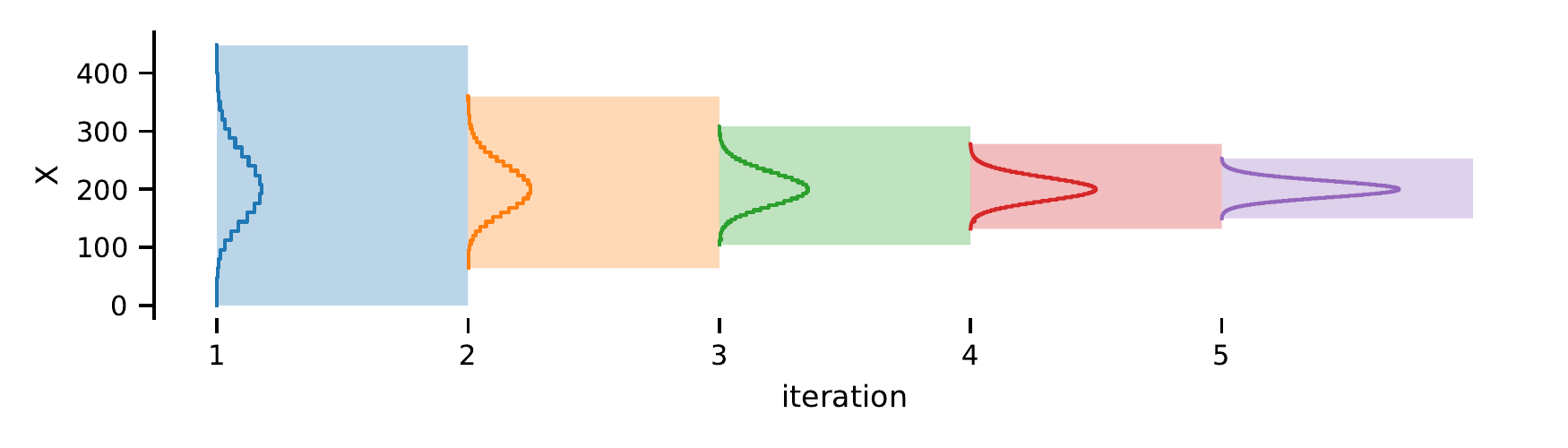}
    \caption{The state-space refinement algorithm on a birth-death process. From left to right the state size is halved and states with low probability are removed from the truncation. The final truncation is a typical truncation with states of size 1 and the initial states are of size $2^4$.}
    \label{fig:refinement}
\end{figure}

\begin{algorithm}
\SetKwFunction{Split}{split}
\SetKwInOut{Input}{input}
\SetKwInOut{Output}{output}
\Input{Initial partitioning $\mathcal{S}^{(0)}$, truncation threshold $\epsilon$}
\Output{approximate stationary distribution $\hat\pi_{\infty}$}
\For{$i=1,\dots,m$}{
    ${\hat\pi}^{(i)}_{\infty}\leftarrow $ solve approximate stationary distribution on $\mathcal{S}^{(i)}$\label{line:stationary}\;
    $\mathcal{R}\leftarrow$ choose smallest $\mathcal{R}'\subseteq\mathcal{S}^{(i)}$ such that $ \sum_{\bar{x}\in\mathcal{R}'}\hat{\pi}_{\infty}^{(i)}(\bar{x})\geq1-\epsilon$\label{line:filter}\;
    $\mathcal{S}^{(i+1)}\leftarrow \bigcup_{\bar{x}\in\mathcal{R}} \text{split}(\bar{x})$\label{line:refine}\;
    update $\hat{Q}$-matrix\label{line:update_q}\;
}
\Return ${\hat\pi}^{(m)}_{\infty}$\;
    \caption{Lumping to approximate the stationary distribution }
    \label{alg:refinement}
\end{algorithm}
The refinement algorithm (Alg.~\ref{alg:refinement}) starts with a set of large macro-states
that are iteratively refined, based on approximate stationary distributions.
We start by constructing square macro-states of size
$2^m$ in each dimension for some $m\in\mathbb{N}$ such that they form a large-scale grid $\mathcal{S}^{(0)}$. 
Hence, each initial macro-state has a volume of ${\left(2^m\right)}^{n_S}$.
This choice of grid size is convenient because we can halve states
in each dimension.
Moreover, this choice ensures that all states have an equal volume
and we end up with unit-sized macro-states,
equivalent to a truncation of the original non-lumped state-space.

An iteration of the state-space refinement starts by computing the stationary distribution, using the lumped $\hat Q$-matrix.
Based on a threshold parameter $\epsilon>0$
states are either removed or split (line~\ref{line:refine}), depending on
the mass assigned to them by the approximate stationary
probabilities $\hat\pi^{(i)}_{\infty}$.
Thus, each macro-state is either split into $2^{n_S}$ new states or removed
entirely.
The result forms the next lumped state-space $\mathcal{S}^{(i+1)}$.
The $\hat{Q}$-matrix is updated (line~\ref{line:update_q}) using \eqref{eq:lumped_q} to calculate the transition rates of the next aggregated truncation $\mathcal{S}^{(i+1)}$. 
Entries of truncated states are removed from the updated transition matrix.
Transitions leading to them are
re-directed according to the re-entry matrix (see Section \ref{sec:fsp}).
After $m$ iterations (we started with states of side lengths $2^m$)
we have a standard finite state projection scheme
on the original model tailored to
computing an approximation of the stationary distribution.

This way, the refinement algorithm focuses only on those parts of the state-space contributing most to the stationary distribution.
For instance, in Fig.~\ref{fig:refinement} the stationary probability mass mostly concentrates around $\#S=200$.
Therefore, states that are further away from this area can be dropped in further refinement.
This filtering (line~\ref{line:filter} in Algorithm~\ref{alg:refinement}) ensures that
states contributing significantly to $\hat\pi_{\infty}^{(i)}$ will be kept and refined in the next iteration.
The selection of states is done by sorting states in descending order according to their approximate probability mass.
This ensures the construction of the smallest possible subset chosen for refinement according to the approximation.
Then states are collected until their overall approximate mass is above $1-\epsilon$.

An interesting feature of   the aggregation scheme is that the distribution
tends to spread out more.
This is due to the assumption of a uniform distribution inside macro-states.
To gain an intuition, consider a macro-state that encompasses a peak of the stationary distribution.
If we re-distribute the actual probability mass inside this macro-state uniformly,
a higher probability is assigned to states at the macro-state's border.
When plugging such macro-states together, this increased mass away from the peak will
increase the mass assigned to adjacent macro-states.
This effect is illustrated by the example of a birth-death process in Figure~\ref{fig:refinement}.
Due to this effect, an iterative refinement typically keeps an over-approximation in terms of state-space area.
This is a desirable feature since relevant regions are less likely to be pruned due to lumping approximations.

\section{Results}\label{sec:results}
A prototype was implemented in Rust~1.50 and Python~3.8.
The linear systems were solved either using Numpy~\cite{numpy} for up to 5000 states, or the sparse linear solver as available through Scipy~\cite{2020SciPy-NMeth}, or the iterative biconjugate gradient stabilized algorithm~\cite{van1992bi} (up to $10,\!000$ iterations and absolute tolerance $10^{-16}$).

The examples that we consider in the sequel 
are typical benchmarks for the analysis of MPMs. For most of them, appropriate Lyapunov functions
have been determined using Geobound \cite{spieler2014numerical}.
However, the corresponding Lyapunov sets containing at least $1-\epsilon_{\ell}$ of the stationary probability mass are very large for typical choices of $\epsilon_{\ell}$ (e.g. $\epsilon_{\ell}\in \{0.1,0.05,0.001\}$). Even
for extremely large $\epsilon_{\ell}$, say $\epsilon_{\ell}=0.8$, the remaining state-space may still be huge (e.g, 15,198 states).
\subsection{Parallel Birth-Death Process}
We first examine the algorithm on the simple example of two parallel, uncoupled birth-death processes.
\begin{model}[Parallel Birth-Death Process]\label{model:par_bd}
Two uncoupled parallel birth-death processes result in a simple stationary distribution that is
given by a product of two Poisson distributions.
$$\varnothing\xrightarrow{\rho} A \qquad A\xrightarrow{\delta} \varnothing \qquad
\varnothing\xrightarrow{\rho} B \qquad B\xrightarrow{\delta} \varnothing$$
As a parameterization we choose $\rho = 100$ and $\delta=1$.
\end{model}
For this model, the stationary distribution is known to be the product of two Poisson distributions with rate $\rho / \delta$.

\begin{figure}[t]
    \centering
    \begin{minipage}{.48\textwidth}
    \centering
    \includegraphics[scale=0.5]{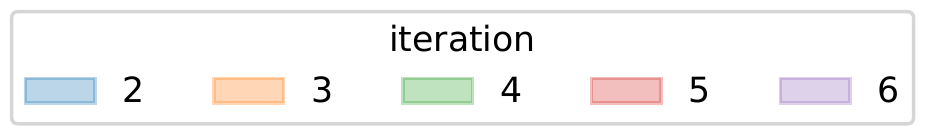}\\
    \vspace{-5mm}
    \includegraphics[width=\textwidth]{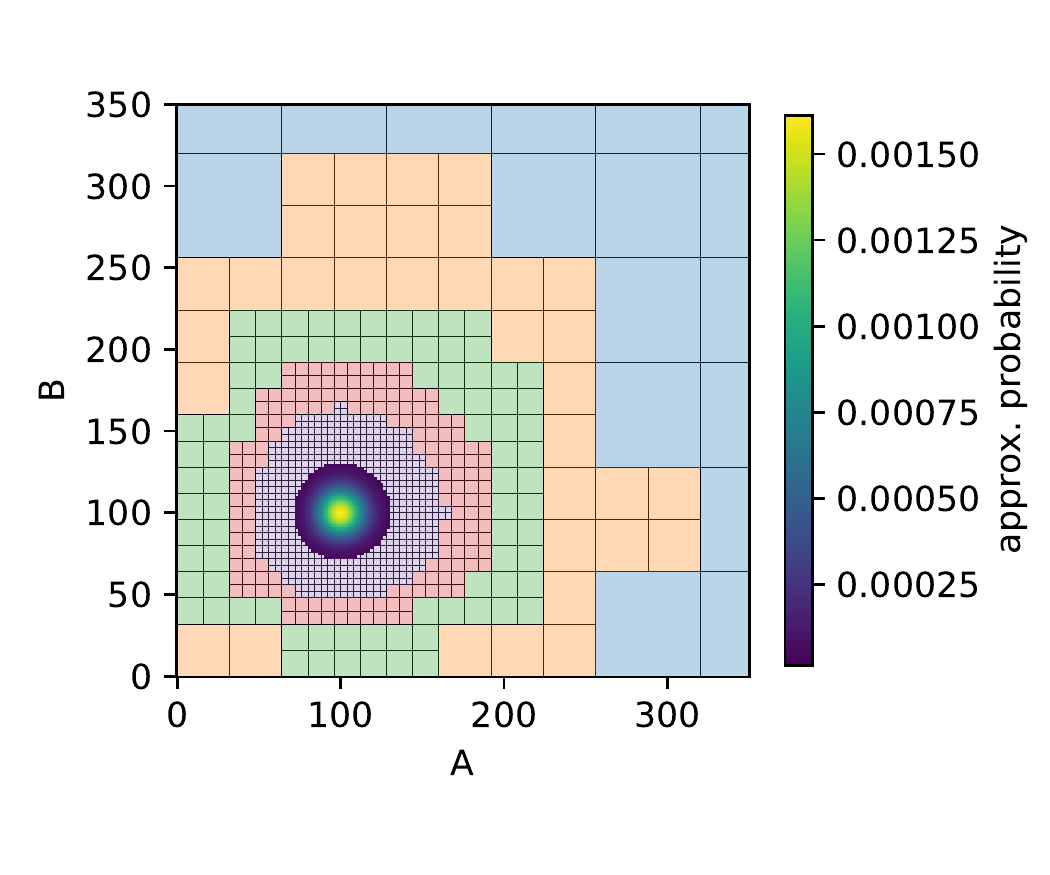}
    \end{minipage}
    \begin{minipage}{.43\textwidth}
    \includegraphics[width=\textwidth]{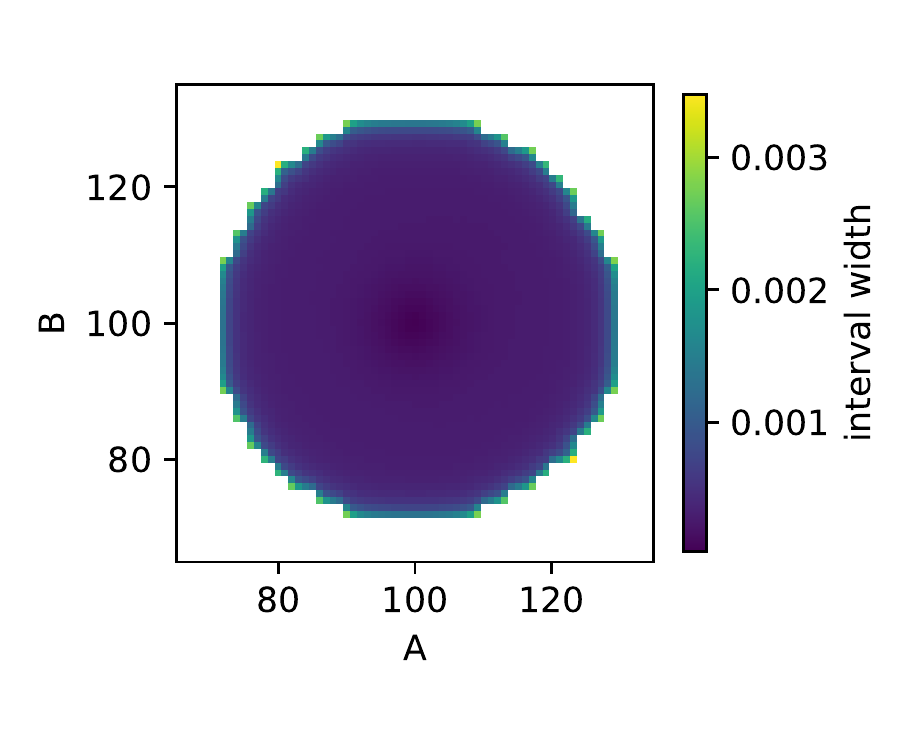}
    \end{minipage}
    \caption{Results for Model~\ref{model:par_bd} with truncation threshold $\epsilon=0.1$. (left) Truncations of different iterations are layered on top of each other. At higher iterations, truncations cover less area but increase in detail, due to the refinement of macro-states. The final approximation is indicated by its approximate probabilities. (right) The difference between the upper and lower bounds on the probability conditioned on the truncation.}
    \label{fig:par_bd}
\end{figure}

According to the Lyapunov analysis with a 1e-4 bound, we fix the initial truncation to a $70\times 70$ grid of macro-states with size $2^7$ in each dimension.
This implies 8 iterations of the algorithm to arrive at a truncation with the original granularity.
In Figure~\ref{fig:par_bd}, we illustrate the truncations of different iterations.
Over the iterations, the covered area decreases, while the aggregation granularity increases.
The final truncation distribution approximation is also depicted and covers $1 - \text{1.27e-2}$ of the true stationary distribution (cf.\ Table~\ref{tab:intervals}).

For this case study, we also compute state-wise bounds on the probabilities conditioned on the truncation as discussed in Section~\ref{sec:fsp}.
In Figure~\ref{fig:par_bd_errors} (right), we present the difference between upper and lower bound for $\epsilon=0.1$.
We observe intervals that are narrowest in the truncation's interior near the distribution's mode.
The largest intervals or the largest absolute uncertainty is present in the boundary states.
This indicates, that the specific reentry distribution has little effect on the main approximate stationary mass.
More detailed results on the intervals' magnitudes are given in Table~\ref{tab:intervals}.

\subsection{Exclusive Switch}
The exclusive switch~\cite{barzel2008calculation}  has three different modes of operation, depending on the
DNA state, i.e.\ on whether a protein of type one or two is bound to
the DNA.

\begin{model}[Exclusive Switch]\label{model:excl_switch}
The exclusive switch model consists of a promoter region
that can express both proteins $P_1$ and $P_2$. Both can bind to the region, suppressing
the expression of the other protein. For certain parameterizations, this leads to a
bi-modal or even tri-modal behavior.
$$ D \xrightarrow{\rho_1} D + P_1 \qquad D \xrightarrow{\rho_2} D + P_2 \qquad P_1 \xrightarrow{\lambda}\varnothing \qquad P_2 \xrightarrow{\lambda} \varnothing $$
$$ D + P_1 \xrightarrow{\beta} D.P_1 \qquad D.P_1 \xrightarrow{\gamma_1} D + P_1 \qquad D.P_1 \xrightarrow{\rho_1} D.P_1 + P_1 $$
$$ D + P_2 \xrightarrow{\beta} D.P_2 \qquad D.P_2 \xrightarrow{\gamma_2} D + P_2 \qquad D.P_2 \xrightarrow{\rho_2} D.P_2 + P_2 $$
We choose parameter values $\rho_1 = 0.7$, $\rho_2 = 0.6$, $\lambda=0.02$, $\beta=0.005$, $\gamma_1 = 0.06$, and $\gamma_2 = 0.05$.
\end{model}
Since the exclusive switch models mutually exclusive binding of proteins to a single genetic locus,
we know a priori that there are exactly three distinct operating modes.
In particular are $D$, $D.P_1$, and $D.P_2$ mutually exclusive such that $X_{D}(t) + X_{D.P_1}(t) + X_{D.P_2}(t) = 1$, $\forall t\geq 0$.
This model characteristic often leads to bi-modal stationary distributions, where one or the other protein is more abundant depending on the genetic state.

Accordingly, we adjust
the initial truncation:
The state-space for the DNA states is not lumped. Instead we ``stack''
lumped approximations of the $P_1$-$P_2$ plane upon each other.
Such special treatment of DNA states is common for such models \cite{lapin2011shave}.
Using Lyapunov analysis for threshold $0.001$, we fix an initial state-space of $63\times 63$ macro-states with size $2^7$. Detailed results for different parameters $\epsilon$ are presented in Table \ref{tab:excl_switch}.
We compute error bounds using a worst-case analysis based on reference solutions provided by Geobound with $\epsilon_{\ell}=0.01$.
We observe a strong decrease in both upper bounds on the total absolute and maximal absolute error in the final iteration.
Interestingly, the errors between different thresholds are very close in earlier iterations.
This is mainly due to the usage of absolute errors which causes probabilities close to the mode dominate.

Using Geobound we observe that our final truncation captures the stationary mass very well (cf.\ Table~\ref{tab:intervals}).
We use the Geobound's lower bounds with $\epsilon_{\ell}=1e-2$ and find that the uncovered mass by the aggregation-based truncation is magnitudes lower than $\epsilon$ or close to it (for $\epsilon=0.1$).
While they capture the mass well, they are much smaller than the Geobound truncation ($\epsilon_{\ell}=0.1$) 
with 16,780 states, regardless of the threshold parameter $\epsilon$.

\begin{figure}
    \centering
    \includegraphics[width=0.39\textwidth]{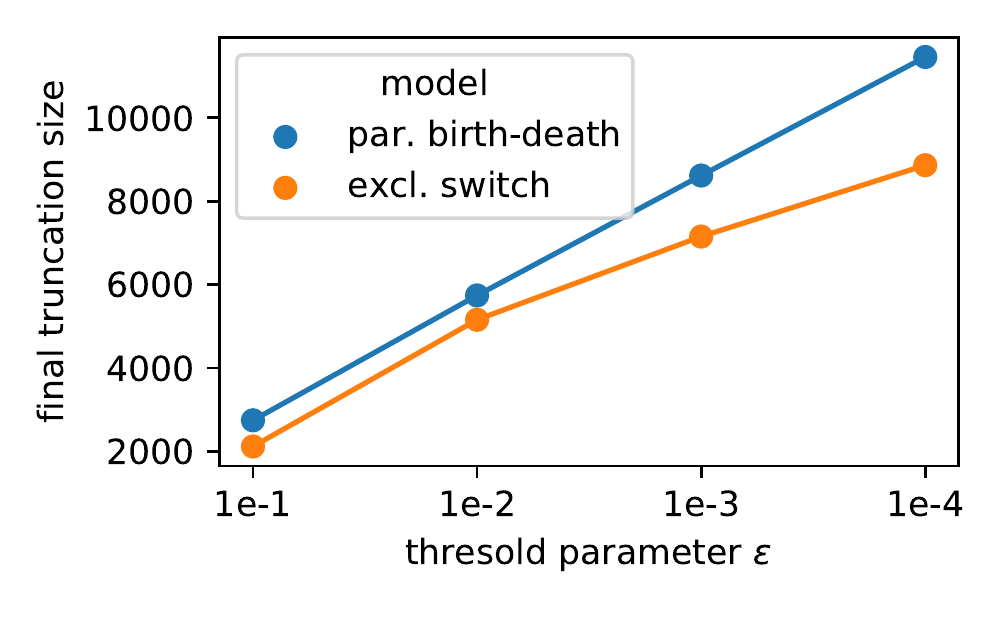}
    \includegraphics[width=0.57\textwidth]{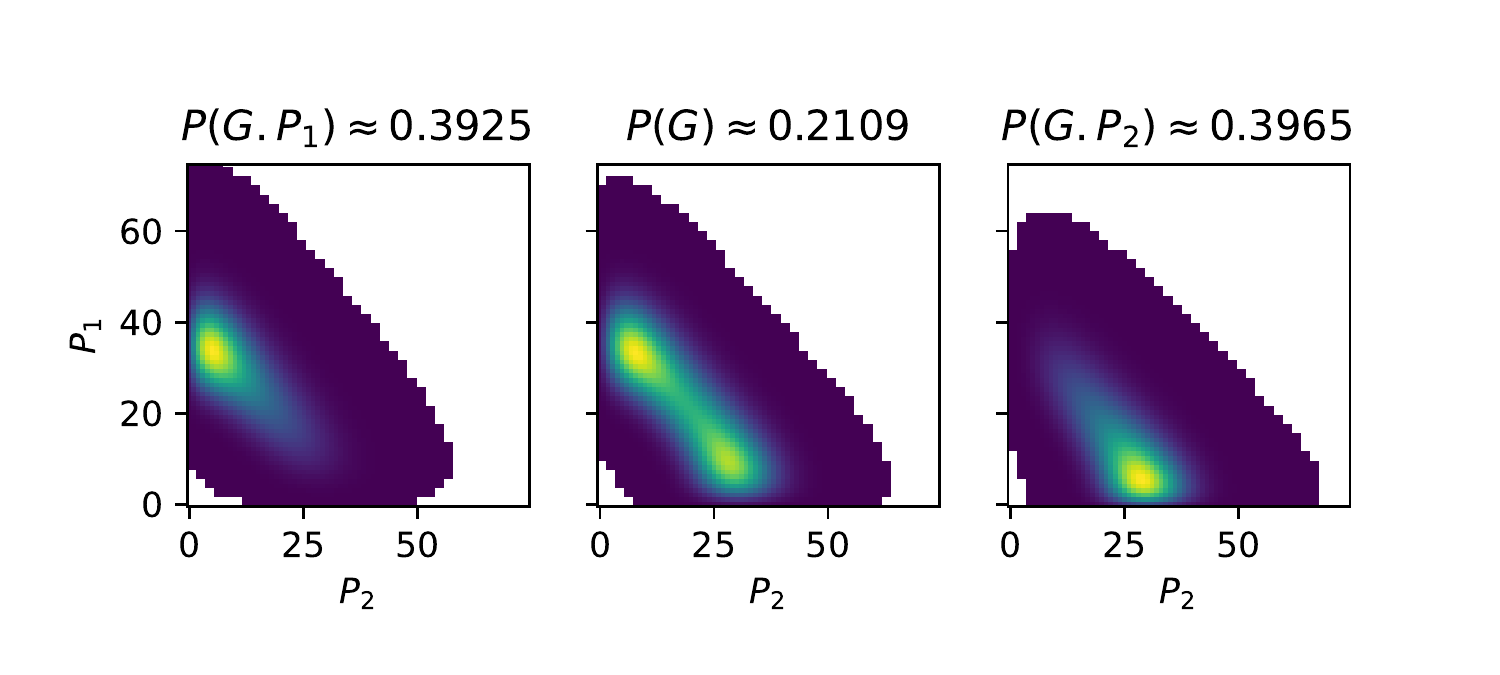}
    \caption{(left) The sizes of the final truncation vs.\ the threshold parameter $\epsilon$. (right) The approximate stationary distribution of the exclusive switch (Model~\ref{model:excl_switch}) obtained with $\epsilon=\text{1e-4}$.}
    \label{fig:excl_switch}
\end{figure}

In Figure~\ref{fig:excl_switch} (left), we show the effect of the threshold parameter $\epsilon$ on the size of the final truncation.
We observe a roughly linear increase in size with an exponential decrease of $\epsilon$.
\begin{table}[t]
    \centering
    \begin{tabular}{L{15mm} L{23mm} R{18mm} R{18mm} R{18mm} R{18mm}}
    \toprule
     \multirow{2}*{Model} & & \multicolumn{4}{c}{threshold parameter $\epsilon$} \\\cmidrule(lr){3-6}
     &  & 1e-1 & 1e-2 & 1e-3 & 1e-4 \\
     \midrule
     \multirow{3}*{\ref{model:par_bd}}
         & total width & 1.2336 & 3.0938e-02 & 5.3916e-04 & 8.1249e-06 \\
         & max.\ width &  3.4752e-03 & 9.2954e-05 & 4.0400e-07 & 4.6521e-09 \\
         & outside mass & 1.2708e-02 & 1.0568e-04 & 1.0500e-06 & 1.0617e-08 \\
         \midrule
     \multirow{3}*{\ref{model:excl_switch}}
         & total width & 5.5171 & 1.5559 & 2.8946e-02 & 3.7161e-04 \\
         & max.\ width & 1.5898e-01 & 3.3089e-03 & 3.4733e-05 & 3.8412e-07 \\
         & outside mass $\leq$ & 1.5274e-01 & 1.2973e-03 & 2.0249e-05 & 2.7280e-07 \\
         \bottomrule
    \end{tabular}
    \caption{Results for Model~\ref{model:par_bd} and Model~\ref{model:excl_switch}: The characteristics of the lower-upper bound intervals on the conditional probability and the (upper bound on) mass not contained in the truncation are given.}
    \label{tab:intervals}
\end{table}

\subsection{p53 Oscillator}
We now consider a model of the interactions of the tumor suppressor p53~\cite{geva2006oscillations}. The system describes the negative feedback loop between  p53 and the oncogene Mdm2.
Species pMdm2 models a precursor to Mdm2. This model is particularly interesting due to its complex three-dimensional oscillatory behavior.
The model is ergodic with a unique stationary distribution~\cite{gupta2014scalable}.
\begin{model}[p53 Oscillator]\label{model:p53}
\begin{align*}
\varnothing \xrightarrow{k_1} \mathrm{p53} \qquad
\mathrm{p53} \xrightarrow{k_2} \varnothing \qquad
\mathrm{p53} \xrightarrow{k_4} \mathrm{p53} + \mathrm{pMdm2}
\\
\mathrm{p53} \xrightarrow{\alpha_4(\cdot)} \varnothing \qquad
\mathrm{pMdm2} \xrightarrow{k_5} \mathrm{Mdm2} \qquad
\mathrm{Mdm2} \xrightarrow{k_6} \varnothing
\end{align*}
The non-polynomial degradation reaction rate
$$
\alpha_4(x) =k_3 x_{\mathrm{Mdm2}} \frac{x_{\mathrm{p53}}}{x_{\mathrm{p53}} + k_7}\,.
$$
The parameterization based on \cite{ale2013general} is $k_1=90$, $k_2=0.002$, $k_3=1.7$, $k_4=1.1$, $k_5=0.93$, $k_6=0.96$, and $k_7 = 0.01$.
\end{model}
With the exception of propensity function $\alpha_4$, we can compute the transition rates $\bar{\alpha}_i$ using the Faulhaber formulae, as discussed in Section~\ref{sec:aggregation}.
We consider $\alpha_4$ separately, because it is non-polynomial and therefore, we have to make an approximation.
The fraction occurring in the non-linear propensity function $\alpha_4$ can roughly be characterized as an activation function:
Due to the low value of parameter $k_7=0.01$ we can approximate
$$\frac{x_{\mathrm{p53}}}{x_{\mathrm{p53}} + k_7}
\approx
\begin{cases}
0 & \text{if } x_{\mathrm{p53}} = 0\\
1 & \text{otherwise}
\end{cases}
$$
We use this approximation at the coarser levels of aggregation to efficiently compute the approximate transition rate $\bar{\alpha}_4$.
At the fines granularity we switch back to exact propensity function $\alpha_4$.\footnote{We note, that $\sum_{i=0}^n i / (i + k_7)$ can be solved analytically. However, the approximation presented above is much simpler to compute.}

\begin{figure}
    \centering
    \begin{minipage}{0.4\textwidth}
    \centering
    \includegraphics[width=\textwidth]{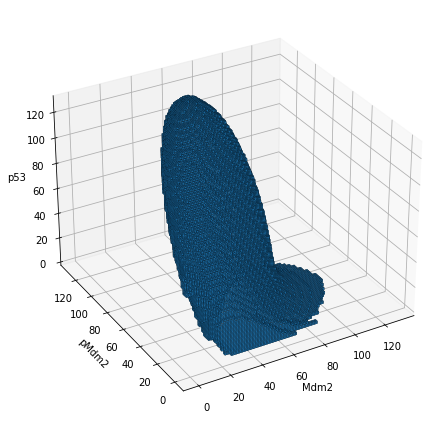}
    \end{minipage}
    \begin{minipage}{0.58\textwidth}
    \centering
    \includegraphics[width=\textwidth]{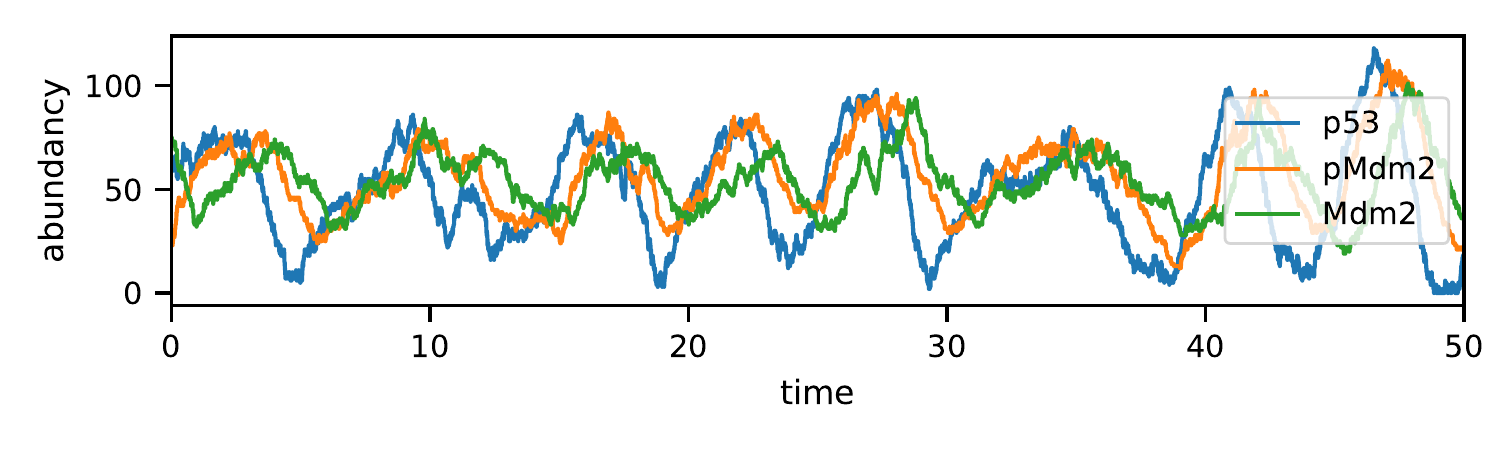}
    \vspace{-10mm}\\
    \includegraphics[width=\textwidth]{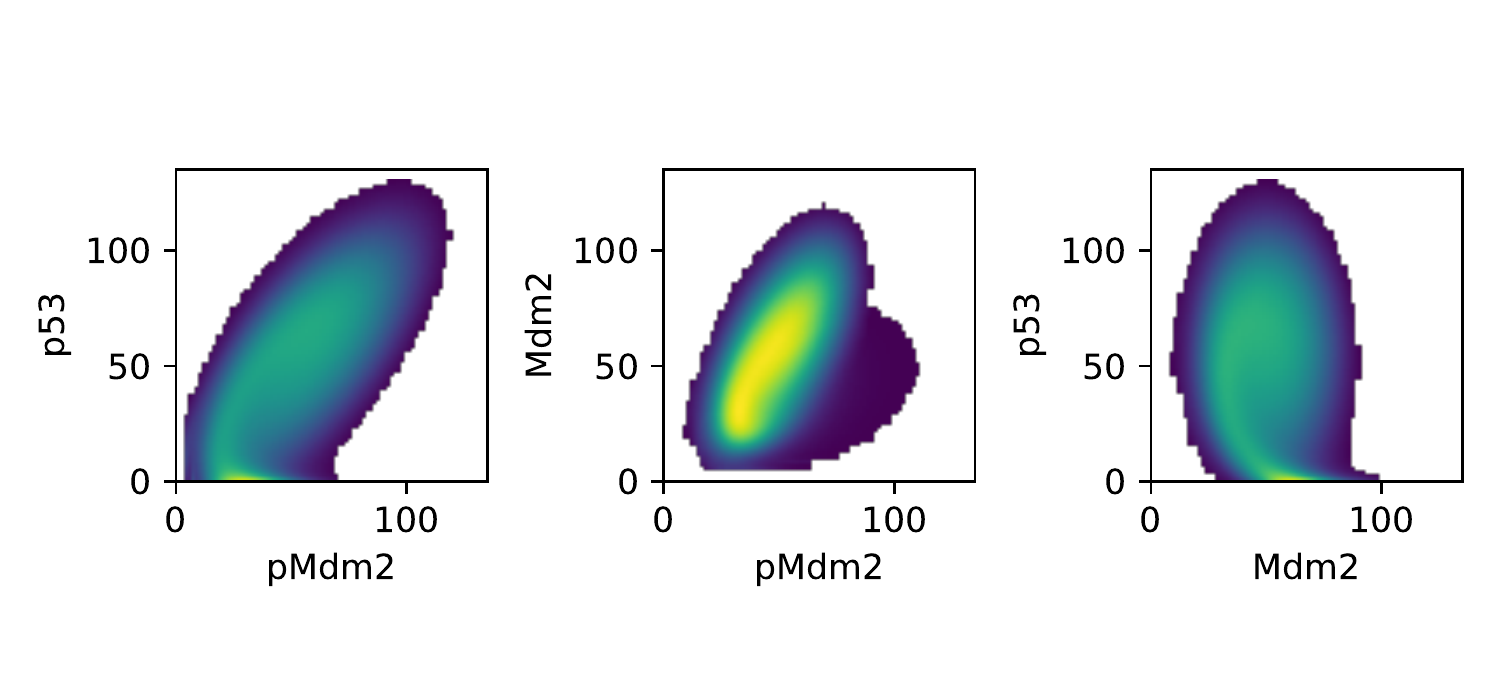}
    \end{minipage}
    \caption{(left) The final truncation at original granularity derived for the p53 oscillator. (top right) A sample trajectory illustrating the oscillatory long-run behavior. (bottom right) The approximate marginal distributions of the stationary distribution based on the truncation derived with $\epsilon=0.1$.}
    \label{fig:p53}
\end{figure}

Due to the exponential increase stemming from the three-dimensional nature of this model, we only evaluated with parameter $\epsilon=0.1$.
According to a Lyapunov analysis (Section~\ref{sec:lyapunov_p53}), the area covered by an $6\times 6\times 6$ macro-states with size $2^{20}$, covers 0.9 of stationary mass.
A truncation of this same area would consist of 226,492,416 states instead of the 216 macro-states.
The model has a striking oscillatory behavior (cf.\ Fig.~\ref{fig:p53} (top right)) that is reflected in its stationary distribution.
This feature is well-captured in the approximate distribution, where the oscillatory behavior leads to a complex stationary distribution (cf.\ Fig.~\ref{fig:p53} (bottom right)).
This distribution leads to a non-trivial truncation (357,488 states) which is tailored to the main stationary mass (Figure~\ref{fig:p53} (left)).

\section{Conclusion}
State-of-the-art methods for numerically calculating the stationary distribution of Markov Population Models rely on coarse truncations of irrelevant parts of large or infinite discrete state-spaces. 
These truncations are either obtained from the stationary statistical moments of the process or from
Lyapunov theory. They are limited in shape because these methods do not take into account the detailed 
steady-state flow within the truncated state-space but only consider the average drift or stationary moments.

Here, we propose a method to find a tight truncation
that is not limited in its shape and iteratively optimizes the set based on numerically cheap solutions
of abstract intermediate models. 
It   captures the main portion of probability mass even in the case of complex behaviors efficiently.
In particular, the method represents another option, where Lyapunov analysis leads to forbiddingly large truncations.

\bibliographystyle{splncs04}
\bibliography{paper.bib}

\appendix
\newpage
\section{Detailed Results}
\begin{table}[h]
    \centering
    \begin{tabular}{L{8mm} l rrrrrrrr}
        \toprule
        \multicolumn{2}{c}{} & \multicolumn{8}{c}{iteration $i$} \\\cmidrule(lr){3-10}
         $\epsilon$ & & 1 & 2 & 3 & 4 & 5 & 6 & 7 & 8  \\
         \midrule
         \multirow{3}*{1e-1} 
         & $|\mathcal{S}^{(i)}|$ & 4,900 & 28 & 52 & 112 & 232 & 472 & 960 & 1,932 \\
         & tot.\ error & 1.91 & 1.84 & 1.73 & 1.55 & 1.29 & 9.35e-1 & 4.88e-1 & 3.54e-2\\
         & max.\ error & 3.15e-3 & 3.13e-3 & 3.08e-3 & 2.98e-3 & 2.77e-3 & 2.38e-3 & 1.57e-3 & 6.04e-5 \\
         \midrule
         \multirow{3}*{1e-2}
         & $|\mathcal{S}^{(i)}|$ & 4,900 & 52 & 104 & 208 & 464 & 988 & 2,008 & 4,052 \\
         & tot.\ error & 1.91 & 1.84 & 1.73 & 1.56 & 1.30 & 9.46e-1 & 5.01e-1 & 6.22e-4\\
         & max.\ error & 3.15e-3 & 3.13e-3 & 3.08e-3 & 2.98e-3 & 2.78e-3 & 2.39e-3 & 1.59e-3 & 8.33e-7\\
         \midrule
         \multirow{3}*{1e-3}
         & $|\mathcal{S}^{(i)}|$ & 4,900 & 84 & 152 & 300 & 652 & 1,440 & 2,996 & 6,068 \\
         & tot.\ error & 1.91 & 1.83 & 1.73 & 1.56 & 1.30 & 9.46e-1 & 5.01e-1 & 9.83e-6\\
         & max.\ error & 3.15e-3 & 3.13e-3 & 3.08e-3 & 2.98e-3 & 2.78e-3 & 2.39e-3 & 1.59e-3 & 1.14e-8\\
         \midrule
         \multirow{3}*{1e-4}
         & $|\mathcal{S}^{(i)}|$ & 4,900 & 116 & 212 & 400 & 848 & 1,872 & 3,960 & 8,060 \\
         & tot.\ error & 1.91 & 1.83 & 1.73 & 1.56 & 1.30 & 9.46e-1 & 5.01e-1 & 9.83e-6\\
         & max.\ error & 3.15e-3 & 3.13e-3 & 3.08e-3 & 2.98e-3 & 2.78e-3 & 2.39e-3 & 1.59e-3 & 1.83e-10\\
         \bottomrule
    \end{tabular}
    \caption{Detailed results for Model~\ref{model:par_bd}. The errors are computed wrt.\ the reference Poissonian product. The total absolute error and the maximum absolute errors are given.}
    \label{tab:par_bd}
\end{table}
\begin{figure}
    \centering
    \hfill
    \includegraphics[width=0.24\textwidth]{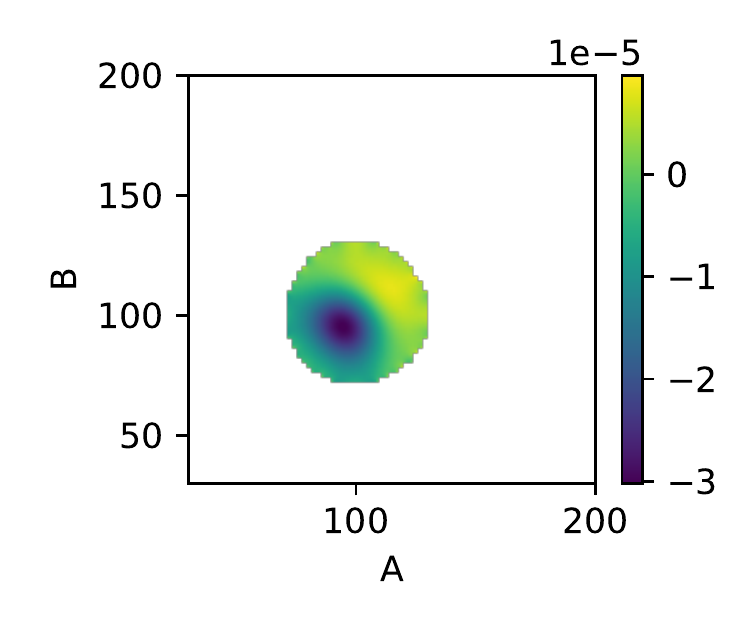}
    \includegraphics[width=0.24\textwidth]{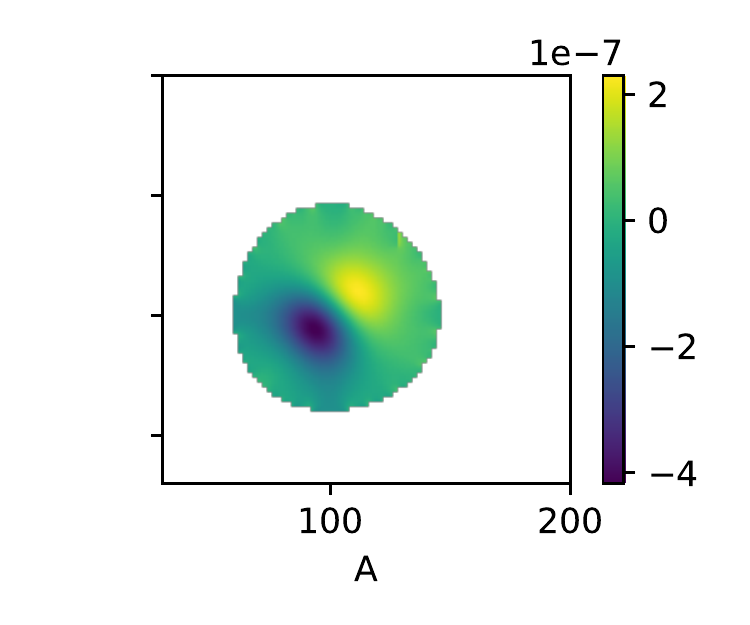}
    \includegraphics[width=0.24\textwidth]{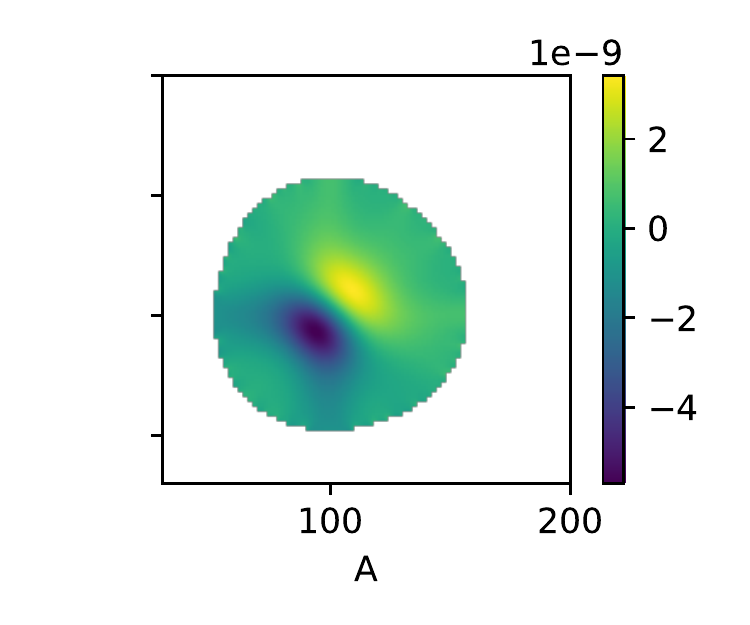}
    \includegraphics[width=0.24\textwidth]{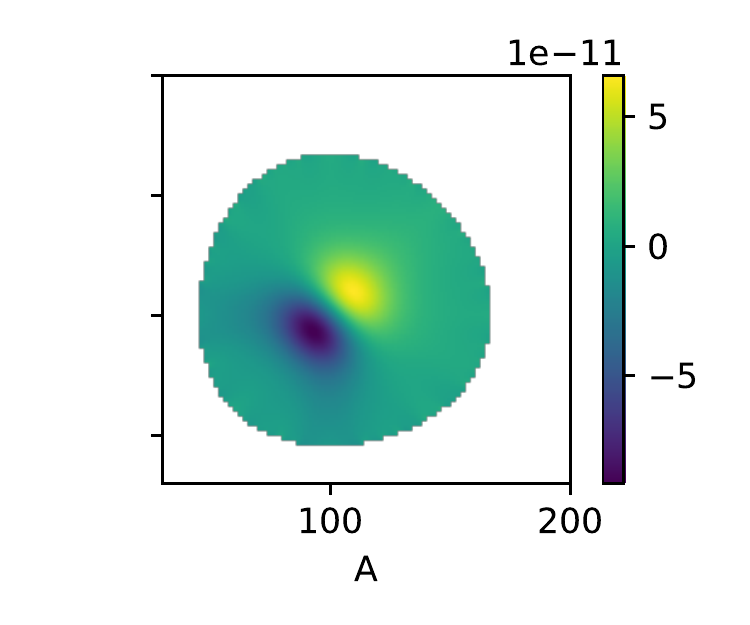}
    \hfill
    \caption{The error over the truncation wrt.\ the analytical solution}
    \label{fig:par_bd_errors}
\end{figure}
\begin{table}
    \centering
    \begin{tabular}{L{8mm} l rrrrrrrr}
        \toprule
        \multicolumn{2}{c}{} & \multicolumn{8}{c}{iteration $i$} \\\cmidrule(lr){3-10}
         $\epsilon$ & & 1 & 2 & 3 & 4 & 5 & 6 & 7 & 8  \\
         \midrule
         \multirow{3}*{1e-1} 
            & $|\mathcal{S}^{(i)}|$ & 11907 & 20 & 32 & 60 & 140 & 340 & 840 & 2116 \\
            & tot.\ error $\leq$ & 1.86e0 & 1.85e0 & 1.45e0 & 1.18e0 & 9.31e-1 & 6.41e-1 & 4.67e-1 & 4.89e-1 \\
            & max.\ error $\leq$ & 1.63e-3 & 1.63e-3 & 1.55e-3 & 1.40e-3 & 1.22e-3 & 9.36e-4 & 8.40e-4 & 1.40e-3 \\
         \midrule
         \multirow{3}*{1e-2}
& $|\mathcal{S}^{(i)}|$ & 11907 & 48 & 112 & 148 & 300 & 720 & 1892 & 5156 \\
& tot.\ error $\leq$ & 1.86e0 & 1.84e0 & 1.44e0 & 1.21e0 & 9.56e-1 & 6.65e-1 & 3.41e-1 & 3.31e-2 \\
& max.\ error $\leq$ & 1.63e-3 & 1.62e-3 & 1.53e-3 & 1.39e-3 & 1.20e-3 & 9.59e-4 & 5.86e-4 & 5.37e-5 \\
         \midrule
         \multirow{3}*{1e-3}
& $|\mathcal{S}^{(i)}|$ & 11907 & 84 & 192 & 244 & 488 & 1084 & 2692 & 7152 \\
& tot.\ error $\leq$ & 1.86e0 & 1.83e0 & 1.46e0 & 1.22e0 & 9.63e-1 & 6.67e-1 & 3.37e-1 & 8.01e-4 \\
& max.\ error $\leq$ & 1.63e-3 & 2.95e-2 & 1.54e-3 & 1.39e-3 & 1.20e-3 & 9.51e-4 & 5.79e-4 & 1.09e-6 \\
         \midrule
         \multirow{3}*{1e-4}
& $|\mathcal{S}^{(i)}|$ & 11907 & 124 & 324 & 352 & 672 & 1436 & 3408 & 8864 \\
& tot.\ error $\leq$ & 1.86e0 & 1.83e0 & 1.46e0 & 1.22e0 & 9.63e-1 & 6.67e-1 & 3.37e-1 & 1.12e-5 \\
& max.\ error $\leq$ & 1.63e-3 & 3.19e-2 & 1.54e-3 & 1.39e-3 & 1.20e-3 & 9.51e-4 & 5.79e-4 & 1.28e-8 \\
         \bottomrule
    \end{tabular}
    \caption{Detailed results for Model~\ref{model:excl_switch}. Upper bounds on the total absolute error and the maximum absolute error are given. The worst-case errors are computed wrt.\ the reference Geobound solution with $\epsilon_{\ell}=1e-2$.}
    \label{tab:excl_switch}
\end{table}

\section{Lyapunov Analysis of the p53 Oscillator}\label{sec:lyapunov_p53}
We now derive Lyapunov-sets for the p53 oscillator case study (Model~\ref{model:p53}). Let the Lyapunov function
\begin{equation}
    g(x) = 120 x_{\mathrm{p53}} + 0.2 x_{\mathrm{pMdm2}} + 0.1 x_{\mathrm{Mdm2}}\,.
\end{equation}
Then the drift
\begin{align}
    d(x) = & - \frac{k_3 x_{\mathrm{Mdm2}} x_{\mathrm{p53}}}{x_{\mathrm{p53}} + k_7}
            - 0.1 k_6 x_{\mathrm{Mdm2}}
            + 120 k_1 \nonumber\\
          & - 120 k_2 x_{\mathrm{p53}}
            + 0.2 k_4 x_{\mathrm{p53}}
            - 0.1 k_5 x_{\mathrm{pMdm2}} \nonumber \\
        = & - \frac{204 x_{\mathrm{Mdm2}} x_{\mathrm{p53}}}{x_{\mathrm{p53}} + 0.01}
            - 0.096 x_{\mathrm{Mdm2}}
            - 0.02 x_{\mathrm{p53}} \nonumber\\
            &- 0.0093 x_{\mathrm{pMdm2}}
            + 10800\,. \label{eq:p53_drift}
\end{align}
Clearly, $c = \sup_{x\in{S}} d(x) = 10800$.
In particular, the supremum $c$ is at the origin since all non-constant terms are negative.
The slowest rate of decrease for \eqref{eq:p53_drift} is $x_{\mathrm{p53}}$ with $x_{\mathrm{Mdm2}} = x_{\mathrm{pMdm2}} = 0$.
We are content with a superset of a Lyapunov set \eqref{eq:lyapunov_set} for some threshold $\epsilon_{\ell}$.
Therefore taking \eqref{eq:lyapunov_set}, we can solve the inequality
$$
\frac{\epsilon_{\ell}}{c}(c - 0.02 x_{\mathrm{p53}}) > \epsilon_{\ell} - 1
$$
for $x_{\mathrm{p53}}$ and 
\begin{equation}
    \frac{c}{0.02 \epsilon_{\ell}} < x_{\mathrm{p53}}\,.
\end{equation}
Therefore
\begin{equation}
\pi_{\infty}\left(\left\{x\in\mathcal{S} \mid \frac{c}{0.2\epsilon_{\ell}} < \lVert x \rVert \right\}\right) > 1 - \epsilon_{\ell}\,.
\end{equation}
\end{document}